\pdfoutput=1
\documentclass[11pt]{article}

\usepackage[utf8]{inputenc}
\usepackage{hyperref}
\usepackage{setspace}
\usepackage{palatino}
\usepackage{graphicx}
\usepackage{float}
\usepackage{titling} 
\usepackage{lscape}
\usepackage{subcaption}
\usepackage[a4paper, total={6in, 9.5in}]{geometry}
\usepackage{mathptmx}

\usepackage[square,sort,comma,numbers]{natbib}

\usepackage[american]{babel}

\usepackage{setspace}
\spacing{1.1}

\usepackage{titlesec}
\titlespacing*{\section}{0pt}{0.75\baselineskip}{0.5\baselineskip}

\begin{document}

\title{Applying Personal Knowledge Graphs to Health}

\author{Sola Shirai, Oshani Seneviratne, and Deborah L. McGuinness \\
Rensselaer Polytechnic Institute, Troy, New York, USA
}
\date{}

\maketitle


\section{Introduction}




Knowledge-driven systems for decision-making in health care applications are powerful tools to help provide actionable and explainable insights to patients and practitioners. In such systems, knowledge about the particular patient - current condition, historical ailments, etc.  - is central to enable personalized health care.

An example of such a system for personalized health care is a diet and lifestyle decision-making tool for diabetic patients. This system may utilize knowledge from several domain-specific knowledge graphs (KGs), such as a KG of diabetes health care guidelines from the American Diabetes Association and a KG of food and nutrition such as FoodKG \cite{foodkg}. Knowledge about a particular patient is used here to perform context-aware reasoning and personalization of down-stream applications. For example, what the system recommends as a "healthy" meal may differ for among patients based on personal aspects like their current weight or exercise habits.

To facilitate reasoning and decision-making based on personal context, such systems can benefit from integrating personal knowledge about the patient. This extended abstract presents a brief review of existing work surrounding the concept of personal knowledge graphs (PKG), how they could be integrated into personalized healthcare as personal \textit{health} knowledge graphs (PHKG), and the key gaps in existing literature that must be addressed to realize their full potential.

\section{Related Work}

Currently there is limited consensus on what exactly a PKG is. In some works \cite{ghidini_personalized_2019,Safavi2019PersonalizedKG}, the term has been used to express the idea of a personalized \textit{view} or \textit{summarization} of a broader KG. Such an interpretation of the term touches on a relevant aspect of personalization, but it fails to encompass capturing knowledge \textit{about} the person as we desire.

Balog and Kenter \cite{goog_pkg} define a PKG as ``a  source  of structured knowledge about entities and the relation between them, where the entities and the relations between them are of personal, rather than general, importance." Within this definition, the PKG only encapsulates knowledge that is relevant to the particular user. Further details about entities are captured by linking to other KGs. For example, the PKG may contain the entity ``my bicycle," but details about this bicycle are expressed through links to other relevant KGs rather than storing them in the PKG.

Some aspects of this definition for PKGs have been captured in existing work. To collect personal knowledge, \cite{microsoft_pkg} mined personal knowledge from natural language utterances based on Freebase relationships. Their work focused specifically on personally relevant relations, such as place of birth, profession, and family, but the scope of relationships that could be captured was rather limited. A means to access and use knowledge from heterogeneous data sources was developed by \cite{thymeflow_18}. Their system, Thymeflow, collects personal knowledge from sources such as email, contacts, calendar, and location history to allow querying. This knowledge is further enriched through means such as connecting location history, calendar events, and geographic coordinates.

We consider PKGs that capture personal health information that can be used in knowledge driven healthcare to be PHKGs. PHKGs remain a somewhat under-explored topic. A vision for PHKGs applied to treatment of chronic diseases was proposed in \cite{knoesisphkg}, which described some challenges involved in the design and use of PHKGs. Their design ideas include capturing personal data from IoT devices and linking the knowledge about patient health to several disease-specific ontologies, similar to the ontology-driven, personalization methods from \cite{ontology_p}. However, to the best of our knowledge, there is no concrete implementation of PHKGs.

\section{Applying PKGs to Health}


Knowledge-driven systems to support health care decision-making can benefit from using a PKG of health knowledge, or a PHKG. For example, a PHKG might capture knowledge about a patient's allergies, eating habits, and specifics of their diabetes disease and treatments. This knowledge would be integrated with other domain-specific KGs to perform reasoning that is tailored to the particular patient, for example by recommending a meal for the patient based on their allergy restrictions and prescribed carbohydrate intake limit.

We anticipate several benefits to be gained from using PHKGs for personalization in such systems: personal health knowledge, if correctly linked, 
can smoothly integrate into reasoning engines; health-related data from heterogeneous sources should be accessible in a consistent fashion; and PHKGs and the supporting methodology should be reusable in further health care applications that link to new knowledge graphs. However, due to the limited existing work on PKGs and even more sparse literature on PKGs for health, there are still many hurdles left to overcome.

\section{Outstanding Challenges}

We identify the following key challenges that must be addressed in order to develop and utilize PHKGs for personalized, knowledge-driven systems for health care applications.

\textbf{1) Collection and Storage of Personal Health Knowledge:} Personal health knowledge can come from a variety of heterogeneous sources \cite{smartdata}, which requires the development of infrastructure to collect and process. Furthermore, collecting personal medical information (e.g. from electronic health records) will require considerations of privacy and access restriction. A comprehensive methodology of how to structure and store personal health information must also be developed so that new personal health knowledge can be introduced to the PHKG in a consistent manner.

\textbf{2) Linking Personal Health Knowledge to External KGs:} Because the entities captured in a PHKG should only be that which is personally related to a particular user, entities must be linked to external KGs to provide additional knowledge. To help disambiguate, entity linking will likely have to rely on the provenance of where and how the entities were added to the PHKG. Disambiguation challenges may also arise in the presence of conflicting health knowledge (e.g. conflicting information about healthy daily calorie intake) or poorly defined terms (e.g. defining what constitutes a ``healthy" meal). It is also possible that entity linking will be highly application-dependent, which may suggest the importance of developing a standardized \textit{workflow} for defining entity links from the PHKG.

\textbf{3) Maintenance of Personal Health Knowledge:} Once the PHKG has been populated, challenges remain in how to maintain the captured knowledge. Maintenance of PHKGs will involve updating personal knowledge in the PHKG (based on changes detected in external KGs) as well as propagating changes made within the PHKG out to the external KGs. Updating knowledge in one location may lead to inconsistencies among linked knowledge, and how to resolve such conflicts is unclear. Additionally, we must carefully consider how and when to trigger updates: continuously monitoring all changes in all linked KGs is not feasible, but the health condition of an individual may fluctuate frequently. The practical challenges of accessing private information must again be considered for maintenance. PHKG implementations will need to address trade-offs between constantly having the most up-to-date knowledge about a patient versus respecting the patient's personal privacy.

\section{Conclusion}

A great deal of work remains to help realize the emerging paradigm of PHKGs. A variety of research and implementation challenges surrounding the collection, linkage, and maintenance of personal health knowledge must be addressed to leverage PHKGs for personalized, knowledge-driven decision-making tools for health care. Some key research questions that will need to be addressed include how to best develop a structure with which to capture personal health knowledge and how to effectively collect and use personal health knowledge while maintaining privacy.



\section*{Acknowledgments}
This work is partially supported by IBM Research AI through the AI Horizons Network.

\medskip

\bibliography{references} 

\begin{thebibliography}{1}

\bibitem{goog_pkg}
K.~Balog and T.~Kenter.
\newblock Personal knowledge graphs: A research agenda.
\newblock In {\em Proceedings of the ACM SIGIR International Conference on the
  Theory of Information Retrieval (ICTIR)}, 2019.

\bibitem{ghidini_personalized_2019}
A.~L. Gentile, D.~Gruhl, P.~Ristoski, and S.~Welch.
\newblock Personalized {Knowledge} {Graphs} for the {Pharmaceutical} {Domain}.
\newblock In C.~Ghidini, O.~Hartig, M.~Maleshkova, V.~Svátek, I.~Cruz,
  A.~Hogan, J.~Song, M.~Lefrançois, and F.~Gandon, editors, {\em The
  {Semantic} {Web} – {ISWC} 2019}, volume 11779, pages 400--417. Springer
  International Publishing, Cham, 2019.

\bibitem{knoesisphkg}
A.~Gyrard, M.~Gaur, K.~Thirunarayan, A.~P. Sheth, and S.~Shekarpour.
\newblock Personalized health knowledge graph.
\newblock In {\em CKGSemStats@ISWC}, 2018.

\bibitem{foodkg}
S.~Haussmann, O.~Seneviratne, Y.~Chen, Y.~Ne’eman, J.~Codella, C.-H. Chen,
  D.~Mcguinness, and M.~Zaki.
\newblock {\em FoodKG: A Semantics-Driven Knowledge Graph for Food
  Recommendation}, pages 146--162.
\newblock Springer, 10 2019.

\bibitem{microsoft_pkg}
X.~{Li}, G.~{Tur}, D.~{Hakkani-Tür}, and Q.~{Li}.
\newblock Personal knowledge graph population from user utterances in
  conversational understanding.
\newblock In {\em 2014 IEEE Spoken Language Technology Workshop (SLT)}, pages
  224--229, Dec 2014.

\bibitem{thymeflow_18}
D.~Montoya, T.~P. Tanon, S.~Abiteboul, P.~Senellart, and F.~M. Suchanek.
\newblock A knowledge base for personal information management.
\newblock In {\em LDOW@WWW}, 2018.

\bibitem{ontology_p}
D.~Riaño, F.~Real, J.~A. López-Vallverdú, F.~Campana, S.~Ercolani,
  P.~Mecocci, R.~Annicchiarico, and C.~Caltagirone.
\newblock An ontology-based personalization of health-care knowledge to support
  clinical decisions for chronically ill patients.
\newblock {\em Journal of Biomedical Informatics}, 45(3):429 -- 446, 2012.

\bibitem{Safavi2019PersonalizedKG}
T.~Safavi, D.~Mottin, and C.~Belth.
\newblock Personalized knowledge graph summarization : From the cloud to your
  pocket.
\newblock In {\em ICDM}, 2019.

\bibitem{smartdata}
A.~Sheth, U.~Jaimini, K.~Thirunarayan, and T.~Banerjee.
\newblock Augmented personalized health: How smart data with iots and ai is
  about to change healthcare.
\newblock {\em RTSI}, 2017, 09 2017.

\end{thebibliography}

\end{document}